\documentclass[letterpaper, 10 pt, conference]{ieeeconf}  % Comment this line out if you need a4paper

\IEEEoverridecommandlockouts                              % This command is only needed if 
\usepackage{cite}

\usepackage[pdftex]{graphicx}
\DeclareGraphicsExtensions{.pdf,.jpeg,.png}
\usepackage{float}
\usepackage{amsmath}
\usepackage{amsfonts}
\usepackage{tikz}
\usetikzlibrary{shapes,arrows}
\usepackage{booktabs} % tables
\usepackage{multirow}  % multiple rows in tables

\usepackage{array}

\usepackage{url}

\title{\LARGE \bf
Real-time Tracking-by-Detection of Human Motion in RGB-D Camera Networks
}

\author{Alessandro Malaguti$^{1}$, Marco Carraro$^{1}$, Mattia Guidolin$^{2}$, 
Luca Tagliapietra$^{3}$, \\ 
Emanuele Menegatti$^{1}$, and Stefano Ghidoni$^{1}$% <-this % stops a space
\thanks{$^{1}$Alessandro Malaguti, Marco Carraro, Emanuele Menegatti and Stefano Ghidoni are with the Department of Information Engineering, University of Padova,
        Via Gradenigo 6/b 35131 Padova, Italy
        {\tt\small \{emanuele.menegatti, stefano.ghidoni\}@unipd.it}
        }%
\thanks{$^{2}$Mattia Guidolin is with the Department of Management and Engineering, University of Padova,
        Stradella S. Nicola, 3 - 36100 Vicenza, Italy {\tt\small mattia.guidolin@phd.unipd.it}
        }%
\thanks{$^{3}$Luca Tagliapietra is with the Department of Industrial Engineering, University of Padova,
        Via Gradenigo, 6/a - 35131 Padova - 35131 Padova, Italy
        {\tt\small luca.tagliapietra@unipd.it}
        }%
}

\begin{document}

\maketitle
\thispagestyle{empty}
\pagestyle{empty}

%%%%%%%%%%%%%%%%%%%%%%%%%%%%%%%%%%%%%%%%%%%%%%%%%%%%%%%%%%%%%%%%%%%%%%%%%%%%%%%%
\begin{abstract}
This paper presents a novel real-time tracking system capable of improving body pose estimation algorithms in distributed camera networks.
The first stage of our approach introduces a linear Kalman filter operating at the body joints level, used to fuse single-view body poses coming from different detection nodes of the network and to ensure temporal consistency between them.
The second stage, instead, refines the Kalman filter estimates by fitting a hierarchical model of the human body having constrained link sizes in order to ensure the physical consistency of the tracking.
The effectiveness of the proposed approach is demonstrated through a broad experimental validation, performed on a set of sequences whose ground truth references are generated by a commercial marker-based motion capture system.
The obtained results show how the proposed system outperforms the considered state-of-the-art approaches, granting accurate and reliable estimates.
Moreover, the developed methodology constrains neither the number of persons to track, nor the number, position, synchronization, frame-rate, and manufacturer of the RGB-D cameras used.
Finally, the real-time performances of the system are of paramount importance for a large number of real-world applications.
\end{abstract}

%%%%%%%%%%%%%%%%%%%%%%%%%%%%%%%%%%%%%%%%%%%%%%%%%%%%%%%%%%%%%%%%%%%%%%%%%%%%%%%%
\section{Introduction}
\label{sec:intro}

Human Body Pose Estimation (HBPE) is a long-lasting challenge in computer vision. 
The capability to detect and reconstruct the human motion is indeed of paramount importance in many applications, ranging from human movement analysis to human-robot cooperation.
However, despite the high relevance of the topic, the challenge is still far from being effectively and fully addressed. 
This is mainly due to the complexity of tracking in real-time the movements of a highly articulated, self-occluding, three-dimensional, variable system as the human body is.
Furthermore, the goal becomes even more challenging when the requirement is to track multiple subjects in real-time without the aid of any body-mounted external device or marker.
%Human Body Pose Estimation (HBPE) is one of the longest-lasting problems in Computer Vision.
%Indeed, the ability to detect the configuration of a body pose is still a relevant task for two main reasons.
%First, the problem of tracking the motion of a highly articulated, self-occluding, three-dimensional (3D) body is very challenging and, therefore, attractive for the academic community.
%The second reason is the popularity of this topic, due to the abundance of possible applications that can benefit from this technology.

One of the most promising technologies to face this challenge exploits the use of several distributed RGB-D cameras acting as nodes of an extensive heterogeneous network.
The common goal of the approaches relying on this technology is to obtain temporally stable 3D reconstructions of multiple subjects' motion by employing the information coming from the different nodes of the network.

%A common goal, among different approaches based on 
%The objective of this paper is to obtain a temporally stable, full-3D skeletal pose by fusing the information coming from an asynchronous network of RGB-D cameras in real-time.
%A fundamental characteristic of this algorithm is its ability to fuse information from different sensors that do not need to be synchronized or to have the same frame rate.
%When a new detection is computed by one of the cameras, it is used to update the joint positions, irrespective of the frame rate and acquisition time.
%In this way, different RGB-D sensors can be used to build heterogeneous networks following each user's needs and possibilities.

The work presented in this paper addresses this problem by means of a tracking-by-detection approach enhanced with solutions tailored to guarantee temporal and physical consistency to the tracked motion.
%a limb-length optimization step. 
%which exploits the feed of multiple RGB-D cameras.
%is based on \emph{OpenPTrack v2.0}\footnote{www.github.com\/openptrack\/open\_ptrack\_v2}, a Human-Computer Interaction library, which includes a markerless multi-camera body pose estimation system~\cite{carraro2017real}. 
The system exploits the feed of multiple RGB-D cameras placed in the scene: each \textit{detection node} uses a combination of a convolutional neural network together with a depth inference algorithm to 
%implements a state-of-the-art 2D body pose estimator algorithm~\cite{Cao2017RealtimeM2} to 
obtain the single-view 3D pose estimation of all the subjects in the scene.
%Then, the 2D poses are projected in the 3D space using the depth information of the RGB-D camera and
Finally, the single-view poses are fused by the \textit{central processing node} to obtain the final multi-view 3D track of each subject's motion. 
%The focus of this work is on this last part of the overall system, i.e. the tracking of the body joints in the 3D space, given the noisy single-view detections.
A fundamental characteristic of the developed methodology is the capability to fuse each node's detections requiring the network nodes neither to be hard synchronized nor to have the same data production rate.
Indeed, every time a new single-view detection is made available by one \textit{detection node}, the \textit{central processing node} uses it to update the multi-view track ensuring, by construction, the timing consistency.
%A fundamental characteristic of this algorithm is its ability to fuse information from different sensors that do not need to be synchronized or to have the same frame rate.
%When a new detection is computed by one of the cameras, it is used to update the joint positions, irrespective of the frame rate and acquisition time.
%In this way, different RGB-D sensors can be used to build heterogeneous networks following each user's needs and possibilities.

In this paper we propose, in addition to our previous work OpenPTrack\footnote{\texttt{www.github.com/openptrack/open\_ptrack\_v2}}~\cite{carraro2016powerful}, an improved version of the Kalman filter to augment its capability to ensure temporal consistency.
To this end, we developed an adaptation mechanism, similar to the one presented in 
%the context of adaptive and robust Kalman filtering presented in
~\cite{sarkka2015adaptive}, to effectively identify and filter out misleading detections acting as outliers and producing noise and errors on the final 3D tracked motions.
Furthermore, we placed in cascade to this enhanced Kalman filter an optimization mechanism, based on an hierarchical model of the human body, in charge of ensuring the physical consistency of the limb lengths.

%an adaptation mechanism to filter out possible outlier measurements similarly to what has been proposed in the context of adaptive and robust Kalman filtering~\cite{sarkka2015adaptive}. 
%The sequential information generated by the different cameras in the network is first fused by the filter.
%At a second stage, an optimization step, based on the hierarchical model of the human body, takes into account the temporal consistency of the limb lengths.
%The final full-3D skeletal pose is obtained in real-time.

In details, with respect to our previous work ~\cite{carraro2017real}, this paper introduces four novel elements:
%This system enhances our previous work~\cite{carraro2017real} thanks to the introduction of four novel elements: 
(i) a new implementation of the Kalman filter considering in its state all the joint positions and velocities of the skeleton model (Sec.~\ref{subsec:multi_view_skeletal_fusion}),
(ii) a joint confidence feedback to adjust the variance of the measurement noise process of the Kalman filter according to the confidence level associated to each single-view detection (Sec.~\ref{subsec:openpose_confidence}),
(iii) an adaptive scheme to further adjust the variance of the measurement noise process of the Kalman filter when possible outlier detections are found (Sec.~\ref{subsec:outlier}),
(iv) a limb-based optimization mechanism, based on a hierarchical human body model, to ensure the physical consistency of the limb lengths (Sec.~\ref{subsec:skeleton_consistency}).

The accuracy and real-time performances of the developed system have been evaluated on a newly collected dataset.
The dataset includes both static and dynamic movements of up to two healthy subjects recorded simultaneously by our 4 RGB-D camera network and by a state-of-the-art marker-based motion capture system used as ground truth.
%In order to evaluate the performance of our system, we created a new comprehensive test dataset for markerless body pose estimation, exploiting a network composed of 4 cameras.
%The ground truth is obtained with a highly accurate commercially available marker-based motion capture system.
The rest of the paper is organized as follows: Sec.~\ref{sec:soa} analyses the related works and compares them with the proposed approach; Sec.~\ref{sect:system_overview} describes the developed methodology, Sec.~\ref{sec:experiments} reports the results of the performance evaluation, and Sec.~\ref{sec:conclusions} draws the final remarks and considerations.

%\begin{itemize}
%	\item a new Kalman filter-based approach for body pose tracking, that fuses single-view detections from multiple RGB-D cameras. The final full-3D skeletal pose is obtained in real-time introducing temporal consistency on the limb lengths and between the consecutive positions of each joint;
%	
%	\item a comprehensive dataset for indoor markerless motion capture is presented. The dataset features 9 sequences, that are different in the number of actors in the scene and in the type of motion, along with ground truth data obtained with a~very accurate marker-based motion capture system. This dataset is used to obtain a quantitative evaluation of the tracking algorithm performance and compare it with other competing filtering methods.
%\end{itemize}

\section{Related Works}
\label{sec:soa}

A broad range of scientific, industrial, and consumer systems rely on the estimation of the human body pose~\cite{sarafianos20163d}. 
Indeed, this information is needed in several applications, such as action recognition~\cite{han2017simultaneous,zanfir2013moving}, people re-identification~\cite{Ghidoni:2017:MFR:3065975.3066116}, and human-computer interaction~\cite{jaimes2007multimodal}.
Applications in the human-robot interaction field require the robot to closely operate with humans: awareness of the human motion is therefore crucial, both for assisted living~\cite{mccoll2011human,gupta20113d} and for industrial scenarios~\cite{morato2014toward}.
Another class of applications that has strongly boosted the research work is video surveillance~\cite{chen2012we,chen2011combined}, including actions and behaviors recognition of people and crowds to detect abnormalities.
Finally, HBPE can be seen as the main building block for motion capture, i.e. the process of digital reconstructing and analysing people movements.

The capability of providing the body pose estimates at the same time that the actions are performed is indeed a central requirement for the large majority of the aforementioned applications.
In recent years, many research efforts have been spent on obtaining, in real-time, fast and reliable pose estimates~\cite{Cao2017RealtimeM2,mehta2017vnect}, supported by the availability of increasingly powerful computing hardware and sensors, like the first and second generations of Microsoft Kinect (Microsoft Corporation, USA).

%A real-time response is usually needed in security applications, for example, when human actions should be detected in time, or in industrial applications, where human motion is predicted to prevent collisions with robots in shared workspaces~\cite{morato2014toward}. 
%Moreover, the introduction of affordable RGB-Depth sensors, like the first and second generation Microsoft Kinect, has further boosted the research in this field.
The availability of such technologies enabled researchers to develop cost-efficient solutions to real-time human pose and motion tracking ~\cite{carraro2016powerful,zivkovic2010wireless,carraro2016cost}.
To this end, many markerless skeleton detection algorithms were developed~\cite{buys2014adaptable,carraro2016improved}.
However, the employment of a single sensor limits the reliability of the estimates, due to the fact that they are generally affected by occlusions and field-of-view limitations. A common solution seems to be connecting several cameras to form a common network~\cite{carraro2017real, moon2016multiple}.
One of the biggest challenges when exploiting a multiple-camera network consists in the methodology used to merge information from different sensors. Several approaches have been developed to obtain accurate skeleton data in a multi-view environment.
The authors in~\cite{caon2011context} used a multiple-Kinect setup for posture and gesture recognition. They acquired single-view skeletons from each camera and then computed joint coordinates differences.
A similar approach has been used for walking posture assessment in~\cite{kaenchan2013automatic}.
In~\cite{liu2018human}, authors used a distributed RGB-D camera network to feed an information-weighted consensus filter based human pose estimator for activities recognition. In this approach, each sensory node provides a measurement of the target human joint, which is used to update the state of the estimator~\cite{kamal2016distributed}. The main limitations of this system, however, are that it only works offline and is compatible to the tracking of just one subject at a time.
A different method is presented in~\cite{ershadi2018multiple}, where a fully connected pairwise conditional random field is used.
However, these approaches rely only on RGB images, thus not exploiting the depth information provided by each camera.
The work proposed in this paper exploits both RGB and depth data from each camera of the asynchronous network. Furthermore, it implements (i) an improved implementation of the multi-view fusion Kalman filter, (ii) an outlier detection scheme, (iii) a joint confidence adaptation scheme, and (iv) a limb-based optimization step.

\section{System Overview}
\label{sect:system_overview}

\begin{figure}
\begin{center}
\includegraphics[width=0.48\textwidth]{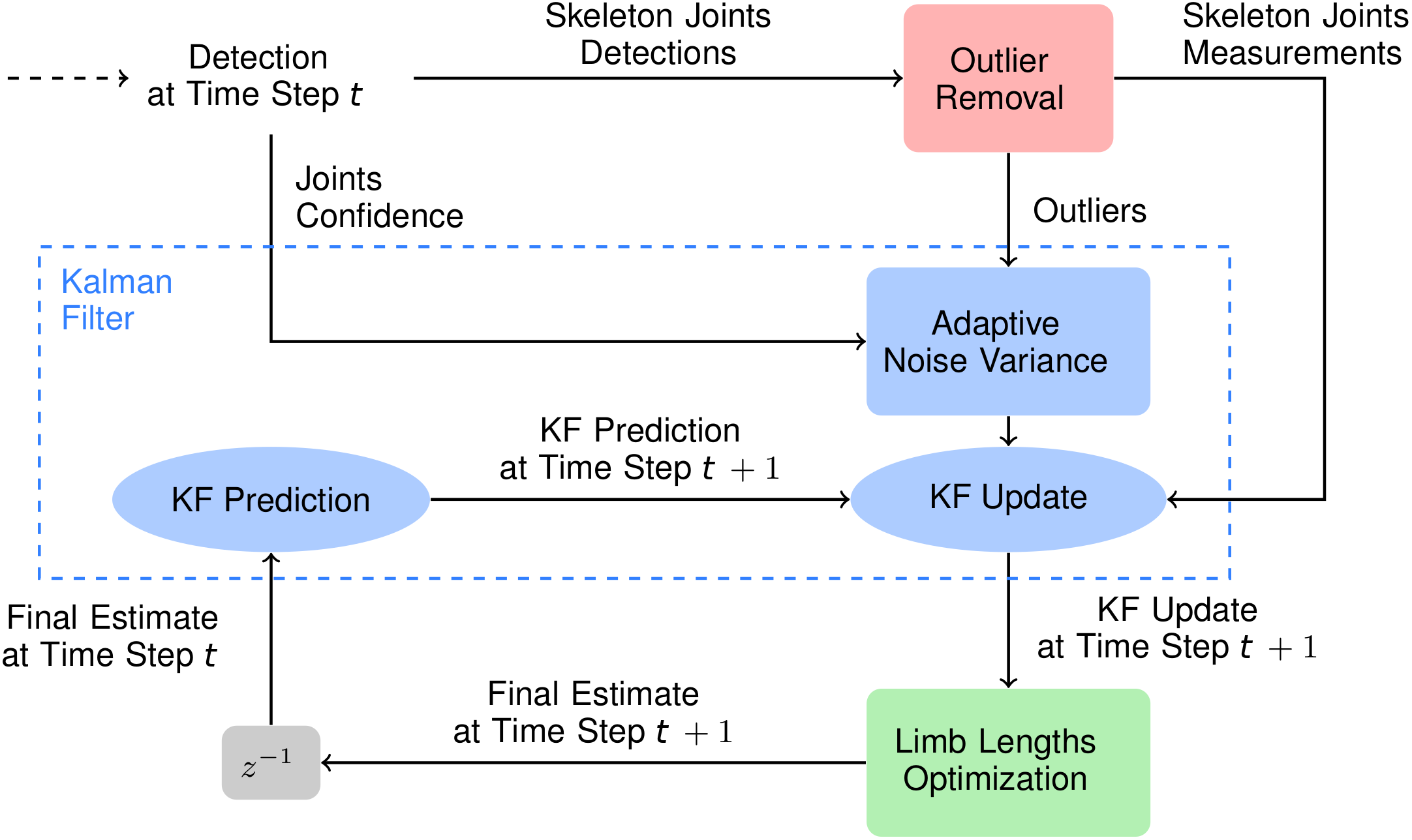}
\end{center}
\caption{Block scheme representation of the overall tracking procedure.\label{fig:tracking_block_scheme}}
\end{figure}

The body pose estimation system relies on the feed of a camera network composed of $N$ asynchronous RGB-D sensors, thus on sequences of RGB images and depth maps from the different sensors of the network.
Neither assumptions on the number of sensor nor on the availability of external aiding tools are made by the system.
%The system does not make any assumption either on the number of sensors, or on the aid of any external marker. 
The only prerequisite is the extrinsic calibration of the camera network\footnote{In this work, we exploited a state-of-the-art open-source approach to compute the extrinsic calibration of the network. See~\cite{munaro2016openptrack} for further information.}.
This calibration procedure firstly defines a common global reference frame $\mathcal{G}$ and then computes the matrix $\mathbf{T}\mathcal{^G_C}_{n}$ expressing the relative transformation between the local frame $\mathcal{C}_{n}$ of the $n$-th camera and the global frame $\mathcal{G}$.

%With this procedure, a common global reference frame $\mathcal{G}$ is defined and a transformation matrix $\mathbf{T}\mathcal{_C^W}$ is obtained for each camera $\mathcal{C}$ in the network. Each of them transforms points from the camera coordinate frame to the world reference frame.
The system can be considered as composed by two parts: (i) the single-view detector and (ii) the multi-view fusion algorithm. 
The first part (i) is the same for each \textit{detection node} of the network (i.e. the computer in charge of acquiring and processing the data coming only from the camera connected to it).
%sensor and is executed in each smart node (i.e., a computer associated with a single sensor)
The multi-view fusion algorithm (ii), instead, is executed only by the \textit{central processing node} which collects and fuses together the estimates provided by the \textit{detection nodes}.

Although a variety of different solutions to estimate 3D body poses from a single RGB-D sensor exists, in this work we used the one described in~\cite{carraro2017real} which extends the single-view approach described in~\cite{Cao2017RealtimeM2}.
Despite our work not being constrained by the specific run-time single-view detection approach, the rationale behind our choice stands in its general applicability since its performances are independent from both the number of people to be tracked and the movements they perform.

\subsection{Single-view detections identification}
The single-view estimates coming from each \textit{detection node} of the network are fused together by the \textit{central processing node}.
At the generic time $t$, it keeps in memory a set of tracks $\mathfrak{T}^\mathcal{G} = \{T^\mathcal{G}_0,\dots,T^\mathcal{G}_{l-1}\}, \quad l \geq 1$, where the generic $T^\mathcal{G}_j \in \mathfrak{T}^\mathcal{G}$ is defined as the vector of the joint positions of the $j$-th skeleton at time $t$, expressed in the global reference frame.
Formally, $T^\mathcal{G}_j = \{\mathbf{g}_m^\mathcal{G} = (x_m^\mathcal{G},y_m^\mathcal{G},z_m^\mathcal{G})\,|\, m =1,\dots,M\}$, where $M$ is the number of joints included in the skeleton.

Let us call $\mathfrak{S}^{\mathcal{C}_n}$ the generic set of skeletons detected by the $n$-th \textit{detection node} in its local reference frame $\mathcal{C}_n$.
Formally, we have 
$\mathfrak{S}^{\mathcal{C}_n} = \{S^{\mathcal{C}_n}_0,\dots,S^{\mathcal{C}_n}_{k-1}\}$, where each $S^{\mathcal{C}_n}_i \in \mathfrak{S}^{\mathcal{C}_n}$ is composed of the position of the $M$ joints of the skeleton.
Whenever the \textit{central processing node} receives a new $\mathfrak{S}^{\mathcal{C}_n}$ from the $n$-th \textit{detection node}, it applies the rigid transformation $\mathbf{T}_{\mathcal{C}_n}^\mathcal{G}$ to express every skeleton in the global reference frame:
%ordered by the model shown in Fig.~\ref{fig:vitruvian},
%is available from the sensor $\mathcal{C}_\matcal{n}$ of the network, the first action to be performed is to transform the 3D joint positions from the camera reference frame $\mathcal{C}$ to the world reference frame:
\begin{equation}
\mathfrak{S}_n^\mathcal{G} = \mathbf{T}_{\mathcal{C}_n}^\mathcal{G} \cdot \mathfrak{S}^{\mathcal{C}_n} = \{ {S_n}^{\mathcal{G}}_j = \mathbf{T}_{\mathcal{C}_n}^\mathcal{G} \cdot {S}^{\mathcal{C}_n}_j, \; \forall {S}^{\mathcal{C}_n}_j \in \mathfrak{S}^{\mathcal{C}_n}\}.
\end{equation}

For the ease of reading, in the following the apex $\mathcal{G}$ will be neglected since every quantity has been already expressed in the global reference frame. Furthermore, also the subscript $n$ will be removed from the notation since, if not explicitly stated, we will consider just the generic $n$-th \textit{detection node}.

The following step performed by the \textit{central processing node} is data association.
The objective is to match each skeleton detected in the current view with its corresponding track, i.e. with its state computed at the previous computation time step. 
If the algorithm fails to find a matching track for one or more of the detected skeletons within the available ones, it generates a new track for each of them, meaning that persons never seen before walked into the scene.

The problem can be mathematically formulated as an assignment problem. To efficiently solve it, we define the cost of the association of track $T_i$ to skeleton $S_j$ and we look for the pair skeleton-track which leads to the minimum total cost~\cite{kuhn1955hungarian, munkres1957algorithms}.
We need therefore to compute, at each time $t$, the assignment cost matrix $C_t$, where its generic element $c_{i,j,t}$ represents the cost of associating the $j$-th skeleton to the $i$-th track.
For each generic $i$-th track among the already available ones, let we call $K_{i}$ the Kalman filter in charge of tracking the position and velocity of its centroid.
We first compute $\hat{z}_{i,t|t-1}$, the Kalman filter prediction for the $i$-th track at time $t$ computed without adding any new detection.
We then compute, for each skeleton detected at time $t$, $z_{i,j,t}$ as the state of $K_{i}$ at time $t$ under the assumption of associating the $j$-th skeleton to the $i$-th track.
%Given the Kalman filter $K_{i}$ that stores and updates the position and velocity of the centroid of the $i$-th track, we define $z_t(i,j)$ as the state of track $i$, if detection $j$ is associated to it, and $\hat{z}_{t|t-1}(i)$ as the Kalman filter prediction of the state of track $i$.
The likelihood vector is therefore computed as:
\begin{equation}
\forall T_i \in \mathfrak{T}, \forall S_i \in \mathfrak{S}: \tilde{z}_{i,j,t} = z_{i,j,t} - \hat{z}_{i,t|t-1}\,.
\end{equation}
Finally, we compute the cost of the $i,j$-th track-skeleton pair as the Mahalanobis distance between $\tilde{z}_{i,j,t}$ and the covariance matrix $\Sigma_{i}$ of the Kalman filter $K_{i}$.
Therefore, each element $c_{i,j,t}$ of the cost matrix $C_t$ uses the Mahalanobis distance, where the covariance matrix of the Kalman filter weights the squared norm of the distance vector:
\begin{equation}
c_{i,j,t} = \tilde{z}^\text{T}_{i,j,t} \cdot \Sigma_i^{-1} \cdot \tilde{z}_{i,j,t}.
\end{equation}
At this point, providing the so-constructed matrix $C_t$ to the Hungarian algorithm~\cite{kuhn1955hungarian}, also known as Munkres algorithm~\cite{munkres1957algorithms}, we solve the assignment problem finding the optimal pairing between tracks and detections.
%The output of the algorithm is therefore the optimal pairing matrix $\mathbf{X}$, where a unit value of one element marks the pairing between the corresponding skeleton and track.
%
%At this point, solving the Hungarian algorithm~\cite{kuhn1955hungarian} associated to the cost matrix $C$ is equivalent to computing the optimal association between tracks and detections.
%In particular, the Munkres algorithm~\cite{munkres1957algorithms} efficiently computes the optimal matrix $\mathbf{X}$, placing a unitary value in the matrix elements corresponding to the associated pairs.
However, since that algorithm does not directly implement a constraint on the maximum distance between tracks and detections, a threshold $\epsilon$ is introduced to reduce the probability of wrong associations.
Given a skeleton, if its distance from all the tracks is higher than $\epsilon$ , than it does not match any of them, meaning that it corresponds to a new person on the scene and, therefore, a new track for him needs to be created.

\subsection{Multi-view skeletal fusion}
\label{subsec:multi_view_skeletal_fusion}
Once the single-view detections identification is completed, the tracks are given as input to the proposed processing pipeline (Fig.~\ref{fig:tracking_block_scheme}). 
%Fig.~\ref{fig:tracking_block_scheme} shows the tracking pipeline developed in the work presented in this paper.
The whole algorithm is based on the association of each track with a $6M$-dimensional Kalman filter, where $M=15$ is the number of joints.
From this point on, for ease of reading, just one track is considered since all the tracks are treated as separate entities and processed following exactly the same pipeline.

The state $\mathbf{x}_t$ of the Kalman filter at time $t$ is constructed by juxtaposition, for each $m$-th joint, of its 3D position $\mathbf{q}_m=[x_m,y_m,z_m]$ and velocity $\mathbf{\dot{q}}_m = [\dot{x}_m,\dot{y}_m,\dot{z}_m]$. Formally:
\begin{equation}
\mathbf{x}_t = \{\left[\mathbf{q}_m \rvert \mathbf{\dot{q}}_m \right]_t^T \,|\, m =1,\dots,M\}.
\end{equation}

%\begin{equation}
%\mathbf{x}_t = \left[x_1,y_1,z_1 \dots x_m,y_m,z_m \rvert \dot{x}_1,\dot{y}_1,\dot{z}_i \dots \dot{x}_m,\dot{y}_m,\dot{z}_m \right]^T.
%\end{equation}
%\begin{equation}
%\mathbf{x}_k = \left[\begin{array}{ccc|ccc} x_1,y_1,z_1 & \dots & x_m,y_m,z_m & \dot{x}_1,\dot{y}_1,\dot{z}_i & \dots & \dot{x}_m,\dot{y}_m,\dot{z}_m \end{array}\right]^T.
%\end{equation}
% SGH - rimosso per limiti di spazio
%The joints are organized following the convention shown in Fig.~\ref{fig:vitruvian}.
The prediction phase of the Kalman filter is driven by the following constant velocity evolution model:
\begin{equation}
\mathbf{x}_{t+1} = \mathbf{F}\mathbf{x}_t + \mathbf{G}\mathbf{n}\,,
\end{equation}
%The matrixes $\mathbf{F}\in \mathbb{R}^{(6m\times6m)}$ and $\mathbf{G}\in \mathbb{R}^{(6m\times3m)}$ are chosen to follow a constant velocity model, since it is good to describe movements in the small temporal window between two good detections of a joint.
%\begin{equation}
%\label{eq:CVmodel}
%\mathbf{F} = \left[
%\begin{array}{ccc|ccc} 1 &  &  & \Delta t & &  \\  & \ddots &  &  & \ddots &  \\  &  & 1 &  &  & \Delta t \\
%\midrule 
% &  &  & 1 &  &  \\  &  &  &  & \ddots &  \\  &  &  &  &  & 1 \end{array}
% \right], \;
%\mathbf{G} = \left[
%\begin{array}{ccc} \frac{\Delta t^2}{2}&  &  \\  & \ddots &  \\  &  & \frac{\Delta t^2}{2} \\ \hline
%\Delta t &  &  \\  & \ddots & \\  &  & \Delta t \end{array}
%\right].
%\end{equation}
where $\mathbf{F} \in \mathbb{R}^{6M\times6M}$ is the transition matrix that implements the constant velocity prediction for each component. 
In other words, at time step $t+1$, the velocity part of the state is predicted to be equal to the previous one ($\mathbf{\dot{q}}_{m,t+1} = \mathbf{\dot{q}}_{m,t}$), while the position part evolves as $\mathbf{q}_{m,t+1} = \mathbf{q}_{m,t} + \Delta t \cdot \mathbf{\dot{q}}_{m,t}$, where $\Delta t$ is the sampling period.
We use $\Delta t = 33\,\text{ms}$, because $30 \, \text{Hz}$ is the maximum frame-rate of our sensor. 
In case of heavy occlusions, a normal situation in multi-persons scenarios, it is likely that the time interval between two consecutive detections of the same full skeleton can be approximated to an integer multiple of $\Delta t$. 
Therefore, letting the time interval between two consecutive detections be $n\cdot\Delta t$, the prediction step is computed $n$ times.
$\mathbf{G} \in \mathbb{R}^{6M\times3M}$ is the noise coupling matrix that describes how the elements of the Gaussian white noise vector $\mathbf{n} \sim \mathcal{N}(\mathbf{0}, \sigma_q^2\cdot\mathbf{1}^{3M\times1})$ affect the system.

The observations at time $t+1$ are represented by the 3D joint positions of the identified skeleton in the global reference frame.
Therefore, the measurement model is:

\begin{equation}
\mathbf{y}_{t+1} = \mathbf{H}\mathbf{x}_{t+1} + \mathbf{w},
\end{equation}
where
\begin{equation}
\renewcommand\arraystretch{1.15}
\mathbf{H} = \left[
\begin{array}{ccc|ccc} 1 &  &  & 0 &  &  \\  & \ddots &  &  & \ddots &  \\  &  & 1 &  &  & 0 \end{array}
\right].
\end{equation}
The meaning of $\mathbf{H} \in \mathbb{R}^{3M\times6M}$ is straightforward, and the measurement noise vector is defined as Gaussian white noise $\mathbf{w} \sim \mathcal{N}(\mathbf{0},  \sigma_r^2\cdot\mathbf{1}^{3M \times 1})$.\\

%\subsection*{Noise variances tuning}
%\label{subsec:parameters_tuning}
Once defined the structure of the Kalman filters, we need to find the values of the noise variances.
In the presented work, we estimated the measurement noise variance offline from a prerecorded static sequence available in our dataset.
We therefore computed the value of $\sigma_r^2$ by averaging, through all the joints, the standard deviation of the joint positions in all the detections of the sequence.
%
%To this end, we computed the standard deviation of the different joint detections, selecting a constant scalar $\sigma_r^2$.
As the careful reader may notice, this is an approximation, given that this value might be variable in different spots of the scene due to a non-perfectly uniform calibration of the camera network.
%is not constant in reality due to the fact that the calibration of the cameras is different in different spots.
%However, this is an estimate useful to capture the order of magnitude of the measurement noise.

The process noise variance, on the other hand, cannot be computed using the same offline procedure since it is dependent from the movements performed by the people in the scene. Therefore, a manual tuning of the process noise variance has been performed.
%On the other hand, the process noise covariance cannot be estimated offline. 
%Indeed, there is no way to determine the uncertainty of the constant velocity motion model beforehand.
%So, a trial-and-error procedure is used to find a value for $\sigma^2_q$ that produces good results.

\subsection{Joint confidence feedback}
\label{subsec:openpose_confidence}

The basic concept of Bayesian filtering is the inclusion of a-priori information in the estimation process.
Starting from this consideration, we preprocess the detections coming from each single-view detection node to gain insights to be included in the system state.
%implemented some pre-processing steps on the output of the joints detections, in order to include information or hypotheses we have on the current state of the system.
The single-view detector returns, for each $m$-th joint of each detection, its 3D position and the associated confidence level $c_m \in [0,1]$.
In order to include this information in our estimation procedure, we implement an adaptive scheme that determines, at time step $t$, the measurement noise variance of $\mathbf{q}_m$, i.e. $\sigma^2_{rc,m_x}, \sigma^2_{rc,m_y}, \sigma^2_{rc,m_z}$ in relation to $c_m$:
\begin{equation}
\label{eq:measurement_noise_variances}
\sigma^2_{rc, m, t} = \sigma^2_{rc,m_x, t} = \sigma^2_{rc,m_y, t} = \sigma^2_{rc,m_z, t} = \frac{\sigma^2_{r}}{c_m}.
\end{equation}
%In this way, when the confidence $c_m$ is close to $1$ the detection is likely to be trusted, as described by $\sigma^2_{r}$, otherwise the current observation's uncertainty is increased inversely proportional to the confidence value.
%Moreover,~\cite{carraro2017real} tries to output a joint detection even in situations with severe occlusions or self-occlusions. 
In order to reduce the tracking errors coming from highly uncertain detections, we use a threshold of $th = 0.5$ to filter them out.
Therefore, if $c_m < th$, the joint detection is rejected and substituted with the Kalman filter prediction at time $t$ ($\hat{y}_{t|t-1}$).
%a substituted for it ($\mathbf{\bar{y}}_{m, t}$) is computed as the weighted average of the last three accepted ones, while the associated noise variance keeps following Eq.\ref{eq:measurement_noise_variances}.
%In order to avoid incorrect detections, we accept a new measure only if the confidence attached to it is superior than a threshold selected as $0.5$. 
%Otherwise, a new measure for joint $j$, necessary for the Kalman filter update, is obtained as a weighted mean of the last three available ones:
%Formally:
%\begin{equation}
%\label{eq:weighted-mean-not-reliable-observation}
%\mathbf{\bar{y}}_{m, t} = \frac{3 \cdot \mathbf{y}_{m, t-1} + 2 \cdot \mathbf{y}_{m, t-2} + 1 \cdot \mathbf{y}_{m, t-3}}{6}.
%\end{equation}
%and an higher noise variance is associated to it. 

\subsection{Outlier filtering}
\label{subsec:outlier}

One of the most important advantages of camera networks is the possibility to overcome occlusions and inconsistencies typical of the single-view detections which generally lead to huge spikes in joint position estimates.
%The objective of this section is to design a way to detect faulty measurements in single-view detections, through the tracking process.
To filter out the so-called outliers, we introduce an adaptive scheme that updates the measurement reliability, based on the recent history of the tracked joint, to prevent too rapid changes in its position. 
%In the literature, the process of finding faulty measurements and dealing with them is often called \emph{outlier rejection}. However, in this thesis we use the term \emph{outlier detection}, in order to highlight the fact that, once a possible outlier has been found, we do not reject the measurement, but we employ an adaptive scheme to make it less reliable.
We consider the Euclidean distance between consecutive positions of the same joint.
Once a threshold is determined, the new measurement is considered unreliable if the distance value from the previous state is above the threshold.
In our algorithm we use, for each $m$-th joint, a slow time-varying threshold $th_m$ that takes into account the information from the joint history.
%Once a joint detection has been determined to be a possible outlier, it is not automatically rejected, but the corresponding measurement noise variance is increased.
%We do not reject the measurements that have been considered as possible outliers, but we update their corresponding noise variance in an adaptive scheme.
The idea is to consider the consecutive distances between the previous $N$ samples of the joint position $d_{m,t-i}$, where $i=1,\dots,N$. The threshold at time $t$ is therefore computed as:
\begin{equation}
th_{m,t} = w\cdot\max\{d_{m,t-i}, \, i=1,\dots,N\},
\end{equation}
where $w\geq1 \in \mathbb{R}$.
\footnote{In our experiments, we found adequate values for $N$ and $w$ to be respectively $15$ and $1.25$.}
In this way, the joint-specific threshold is potentially capable of slowly adapting to the changes in joint speed.

If the new detection distance for the $m$-th joint $d_{m,t}$ is larger than the just computed threshold $th_{m,t}$, the detection is not directly rejected, but the corresponding measurement noise variance, after the joint confidence adaptation, is updated as follow:
%In other words, if the distance is larger than the threshold, the variance will be increased and the measurement will be trusted less and less:
\begin{equation}
\text{if} \,\; d_{m,t} > th_{m,t} \;\; \Rightarrow \;\; \sigma^2_{ro,m,t} = \frac{\sigma^2_{rc,m,t}}{th_{m,t}/d_{m,t}}.
\end{equation}
A possible limitation of this approach might be that correct detections describing a very fast motion of a joint could be detected as outliers. To overcome this limitation we introduced the parameter $o_{max}$, which represents the maximum number of consecutive outliers a track can accept.
In this way, even if $o_{max}$ consecutive measurements are detected as outliers, the next one is considered reliable and used to update the track.
In practice we chose $o_{max} = 2$, based on the idea that, being in a network, if one detection node is experiencing an occlusion leading to an outlier, hopefully the next detection will come from another detection node not experiencing the same occlusion.

\subsection{Skeleton consistency}
\label{subsec:skeleton_consistency}
A typical problem in skeletal tracking from images is the segment length variability. Indeed, the distance between two adjacent joints might vary depending on their relative position estimated from images taken from different viewing angles.
To overcome this limitation, we introduce an algorithm which aims at keeping the segment lengths constant during the whole tracking process.
%Another contribution we have introduced in our tracking algorithm takes into account the lengths of the limbs connecting the skeleton joints, which obviously should remain constant throughout the tracking process.
While the Kalman filter presented in the previous sections focuses on ensuring temporal coherence, this section aims at discussing the algorithm introduced to guarantee the physical consistency of the tracked skeletons.
%between consecutive poses of the human body moving in the 3D space, this skeleton fitting step aims at avoiding limb length inconsistencies and is based on an optimization procedure.

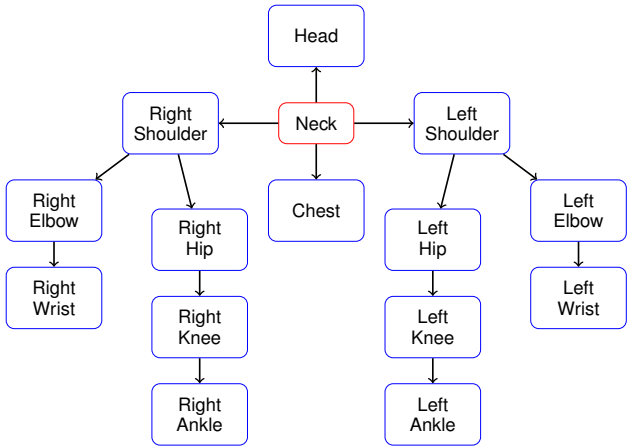
\begin{figure}
	\resizebox{\linewidth}{!}{
	\centering
	%\captionsetup{width=0.75\linewidth}
	\tikzstyle{neck}=[rectangle, draw=red, minimum height=2em,text width=3em,text centered,font = \sffamily\small,rounded corners]
	\tikzstyle{joint} = [rectangle, draw=blue, minimum height=3em,text width=4em, text centered, rounded corners,font = \sffamily\small]
	\begin{tikzpicture}
	\node[neck] (neck) {Neck};
	\node[joint,above of=neck,node distance=1.5cm] (head) {Head};
	\node[joint,below of=neck,node distance=1.5cm] (chest) {Chest};
	\node[joint,right of=neck,node distance=2.5cm] (leftsh) {Left\\Shoulder};
	\node[joint,left of=neck,node distance=2.5cm] (rightsh) {Right\\Shoulder};
	\node[joint,right of=leftsh,node distance=2cm, yshift=-1.5cm] (leftel) {Left\\Elbow};
	\node[joint,left of=rightsh,node distance=2cm, yshift=-1.5cm] (rightel) {Right\\Elbow};
	\node[joint,below of=leftel,node distance=1.5cm] (leftwr) {Left\\Wrist};
	\node[joint,below of=rightel,node distance=1.5cm] (rightwr) {Right\\Wrist};
	\node[joint,below of=leftsh,node distance=2cm,xshift=-0.5cm] (lefthip) {Left\\Hip};
	\node[joint,below of=rightsh,node distance=2cm,xshift=0.5cm] (righthip) {Right\\Hip};
	\node[joint,below of=lefthip,node distance=1.5cm] (leftkn) {Left\\Knee};
	\node[joint,below of=righthip,node distance=1.5cm] (rightkn) {Right\\Knee};
	\node[joint,below of=leftkn,node distance=1.5cm] (leftank) {Left\\Ankle};
	\node[joint,below of=rightkn,node distance=1.5cm] (rightank) {Right\\Ankle};
	\draw[->,thick] (neck) -- (head);
	\draw[->,thick] (neck) -- (chest);
	\draw[->,thick] (neck) -- (leftsh);
	\draw[->,thick] (neck) -- (rightsh);
	\draw[->,thick] (leftsh) -- (leftel);
	\draw[->,thick] (leftel) -- (leftwr);
	\draw[->,thick] (rightsh) -- (rightel);
	\draw[->,thick] (rightel) -- (rightwr);
	\draw[->,thick] (rightsh) -- (righthip);
	\draw[->,thick] (righthip) -- (rightkn);
	\draw[->,thick] (rightkn) -- (rightank);
	\draw[->,thick] (leftsh) -- (lefthip);
	\draw[->,thick] (lefthip) -- (leftkn);
	\draw[->,thick] (leftkn) -- (leftank);
	\end{tikzpicture}
	}
	\caption[Hierarchical model of the human body.]{Hierarchical model of the human body. The nodes represent the joints while the connecting lines the body links.}
	\label{fig:hierarchical_model}
\end{figure}

At each time step $t$, the physical consistency algorithm takes as input the system state $\mathbf{x}_t$, and the hierarchical model of the human body (Fig. \ref{fig:hierarchical_model}).
%The links are defined as the connections between two consecutive nodes in the tree structure represented in Fig. \ref{fig:hierarchical_model}.
Head and chest joints have been excluded from the optimization for two reasons. On the one hand, we obtained more stable results computing the chest as the central point between shoulders and hips. On the other hand, the head has been removed since out of interest for this preliminary assessment. However, the developed implementation of the algorithm already supports the inclusion of those two joints.
At the time $t$ (dependency omitted for the ease of reading), for each $i$-th link of the filtered skeleton track ($i \in 1 \dots L$, with $L$ number of links of the skeleton model), we compute the energy of its length error:
\begin{align}
%\begin{split}
\label{eq:optimization1}
%E_i = (\sqrt{(x-x_p)^2 + (y-y_p)^2 + (z-z_p)^2} - \hat{l}_i)^2,
E_i = (\| \mathbf{q}_{c,i} - \mathbf{q}_{p,i} \| - \hat{l}_i)^2
% \\ i=1,\dots,12,
%\end{split}
\end{align}
where $\mathbf{q}_{c,i}$ and $\mathbf{q}_{p,i}$ are the 3D coordinates of, respectively, the child and parent joints of the $i$-th link. While $\mathbf{q}_{c,i}$ is the output of the current link length optimization, $\mathbf{q}_{p,i}$ is the output of the length optimization of the previous link of the hierarchical model.
Initial values for $\mathbf{q}_{c,i}$ are set equal to the estimates of the Kalman filter $\mathbf{\hat{q}}_{c,i}$.
%where $\mathbf{q}$ is the 3D position of the child joint of the link, $\mathbf{p}_p = (x_p,y_p,z_p)$ is the 3D position of the parent joint already computed and $\hat{l}_i$ is the estimated length of the current link $i$ considered.
%Moreover, initial conditions for $(x,y,z)$ are defined as the positions estimated by the Kalman filter at each time step.

%As already discussed, the overall system does not need skeleton initialization nor we rely on tracking data to estimate the set $\{\hat{l}_i, \; i=1,\dots,12\}$.
We initialize $\hat{l}_i$ for each link using the average of the limb lengths measured in the first few frames in which the entire skeleton is completely tracked.
Then, after the optimization, the estimated lengths are updated using the just computed joint coordinates to slowly compensate possible inaccurate link lengths initializations.
%\footnote{It is important to underline that limb lengths measurements are computed from single-view detections and not from the state estimates. Indeed, a bad initialization of $\{\hat{l}_i, \; i=1,\dots,12\}$ can produce a terrible result throughout the tracking process, if the lengths are updated looking only at the final joints estimates, after they are optimized from the lengths themselves. Using the detections, on the other hand, guarantees a fast recovery from a bad initialization.}.
In order to avoid jitters in the final estimates due to local minima computed by the minimization of the energy defined as in Eq. \ref{eq:optimization1}, we add a second term which accounts for link orientation errors. Therefore, we re-defin $E_i$ as:
\begin{equation}
\label{eq:optimization2}
E_i = (\|\mathbf{q}_{c,i}-\mathbf{q}_{p,i}\|-\hat{l}_i)^2 + (\|\text{\boldmath$\theta$}_{i}-\text{\boldmath$\theta$}_{kf,i}\|)^2
\end{equation}
where:
\begin{equation}
%\begin{align}
\label{eq:orientations}
\text{\boldmath$\theta$}_{i} = \frac{\mathbf{q}_{c,i}-\mathbf{q}_{p,i}}{\|\mathbf{q}_{c,i}-\mathbf{q}_{p,i}\|}\\, \quad
\text{\boldmath$\theta$}_{kf,i} =
\frac{\mathbf{q}_{kf,c,i}-\mathbf{q}_{p,i}}{\|\mathbf{q}_{fk,c,i}-\mathbf{q}_{p,i}\|}\\
%\end{align}
\end{equation}
The unitary vectors $\text{\boldmath$\theta$}_{i}$ and $\text{\boldmath$\theta$}_{kf,i}$ describe, respectively, the orientation of the $i$-th link during the optimization process and the original link orientation estimated by the Kalman filter.
%
%\begin{equation}
%\label{eq:optimization2}
%\begin{aligned}
%\mathbf{x} &= \text{argmin}(E_i), \; i=1,\dots,L, \\ &= \text{argmin} ( w_1\cdot(\|\mathbf{x}-\mathbf{x}_p\|-\hat{l}_i)^2 + w_2\cdot(\|\mathbf{v}-\mathbf{v}_{\text{kf}}\|)^2), \\ & i=1,\dots,L,
%\end{aligned}
%\end{equation}
%\begin{equation}
%\label{eq:optimization3}
%\mathbf{v} = \frac{\mathbf{x}-\mathbf{x}_p}{l}, \; \; \mathbf{v}_{\text{kf}} = \frac{\mathbf{x}_{\text{kf}}-\mathbf{x}_p}{l_{\text{kf}}}.
%\end{equation}
%The vector $\mathbf{v}_{\text{diff}}$ 
%The unit vector $\mathbf{v_i}$ express the 3D orientation of the link, where $l$ is the new link length updated at each iteration of the minimization algorithm.
%On the other hand, $\mathbf{v}_{\text{kf}}$ and $l_{\text{kf}}$ are the constant unit vector and the length between the final Kalman filter estimate of the joint $\mathbf{x}_{\text{kf}}$ and $\mathbf{x}_p$, respectively.
To solve this energy minimization problem we use the Levenberg-Marquardt algorithm implemented in the Ceres Solver library\footnote{\texttt{http://www.ceres-solver.org}}.
%Finally, the actual implementation is done using the  and the .

%\subsection{Overall System Representation}
%
%In this last section dedicated to the details of our algorithm, we present a block scheme representation of the sequence of steps defining the overall tracking procedure.
%
%In this section, we have discussed the skeleton tracking algorithm presented in this thesis.
%In order to obtain a quantitative evaluation of its performance and of the effect of each step of the procedure, we register a dataset, which is described in detail in the following chapter.

\section{Experiments and Results}
\label{sec:experiments}

In order to highlight the efficacy of the proposed approach, given the lack of publicly available RGB-D camera-network datasets, a new one has been recorded.
We are currently working to make our dataset public.
%One of the most tedious tasks in the Computer Vision field is to record datasets and (often manually) annotate the ground truth.
%To speed up the phasesthis process up,
To record the dataset, we asked different subjects to perform free movements in the scene.
We recorded data from a camera network composed by 4 Kinect v2 cameras and, at the same time, the output of a commercial motion capture system (BTS SMART-DX, BTS Bioengineering Corp., USA).
%We registered several trials, at the same time different sequences using a camera network composed by 4 Kinect cameras, together with the output of a commercial MoCap system, the BTS SMART-DX (BTS Bioengineering Corp., Quincy, MA, USA).
%The dataset is composed of a set of multi-view RGB-D sequences where one or two persons performs different free movements on the scene.
Each subject was instrumented with a set of passive reflective markers attached to the skin following the BTS standard marker placing protocol.
%(the MoCap system used has an accuracy of 0.1\,mm) to be used as a reference.
% SGH - rimosso per limiti di spazio
%\footnote{as the careful reader may note, this is needed only for computing the ground truth, since our system does not rely on the markers to estimate its output}. Fig.~\ref{fig:camera_disposition} shows the disposition of the different MoCap cameras and RGB-D sensors in the area used to record.

%\begin{figure}[h!]
%	\centering
%	\captionsetup{width=0.7\linewidth}
%	\includegraphics[width=0.7\textwidth]{lab_cameras}
%	\caption[Disposition of a few of the cameras on the acquisition platform.]{Disposition of a few of the cameras on the acquisition platform: red circles indicate SMART cameras, blue circles indicate Kinect v2 cameras. This picture represents one half of the room, the other one was symmetrical.}
%	\label{fig:camera_disposition}
%\end{figure} 

The 4 Kinects were placed at the four corners of the considered motion area, while the 12 SMART-DX cameras were equally spaced along the perimeter.
% SGH - rimosso per limiti di spazio
%, as shown in Fig. \ref{fig:camera_disposition}.
Both the Kinects and the SMART-DX cameras were placed at a height of approximately 2.5\,m.
% SGH - testo abbreviato per la discussione sui framerate di acquisizione
Noteworthy, the two systems acquire data at different frequencies: 30\,fps for the Kinect sensor network, and 50\,fps for the SMART-DX cameras.
A time synchronization mechanism for relating each Kinect data point with the corresponding reference was therefore set up.

%\begin{figure}[h!]
%	\centering
%	%\captionsetup{width=0.65\linewidth}
%	\resizebox{0.85\linewidth}{!}{
%	\includegraphics[scale=1.0]{figures/camera_disposition_legend.png}	
%%	\begin{tikzpicture}
%%	\node[inner sep=0pt] (image)
%%	{\includegraphics[width=.65\textwidth]{camera_disposition_legend}};	
%%	\node[above of=image,node distance=3.6cm,xshift=-1.425cm,font = \sffamily\small] (smart) {SMART-DX};
%%	\node[above of=image,node distance=3.6cm,xshift=2.5cm,font = \sffamily\small] (kinect) {Kinect v2};
%%	\node[above of=image,node distance=0cm,yshift=-0.5cm,font = \sffamily\small] (platform) {Acquisition Platform};
%%	\end{tikzpicture}
%	}
%	\caption{Outline of the disposition of the cameras around the acquisition platform. The commercial MoCap system used to record the ground data was composed of 12 SMART-DX cameras, while we used 4 Kinect-v2 placed at the corners of the tracking space.}
%	\label{fig:camera_disposition}
%\end{figure}
After completing the calibration procedures of both the systems, we recorded 
%The dataset is composed of a total of 11805 frames, including
three static sequences (15 seconds each) and six dynamic sequences (approx. 60 seconds each).
Each dynamic sequence is characterized by different motion characteristics (i.e. fast or slow movements) performed by a different number of subjects (one or two).
%In the dynamic sequences one or two subjects were present in the scene , that vary in the number of performers (one or two) and the characteristics of the motions (slow or fast).
% SGH - rimosso per limiti di spazio
%, as Table~\ref{tab:sequences} shows.
%\vspace{0.1cm}
%\begin{table}[h]
%	\centering
%	\setlength\tabcolsep{0.5cm}
%	%\captionsetup{width=0.78\linewidth}
%	\def\arraystretch{1.1}
%	\resizebox{\linewidth}{!}{
%	\begin{tabular}{c|ccc}
%		\toprule 
%		\textbf{Sequence} & \textbf{Number} & \textbf{Type} & \textbf{Number}\\ 
%		\textbf{Number} & \textbf{of Actors} & \textbf{of Motion} & \textbf{of Frames}\\ 
%		\midrule
%		\#1 & 1 & slow motions & 1612\\
%		\#2 & 1 & fast motions & 1604\\
%		\#3 & 1 & slow motions & 1503\\
%		\#4 & 1 & fast motions & 1515\\
%		\#5 & 2 & slow motions & 1505\\
%		\#6 & 2 & fast motions & 1517\\
%		\#7 & 1 & static & 861\\
%		\#8 & 1 & static & 749\\
%		\#9 & 2 & static & 939\\
%		\bottomrule
%	\end{tabular}
%	}
%	\caption[Summary of the sequences recorded for the dataset.]{Summary of the sequences recorded for the dataset. The last column reports the number of frames captured by the stereophotogrammetric system at $50\, \text{Hz}$.}
%	\label{tab:sequences}
%\end{table}
While the static sequences have been used for tuning systems parameters, the dynamic ones have been used to assess the accuracy of the proposed approach against the gold-standard results provided by the BTS system.

\begin{figure}
	\centering
	\includegraphics[width=\linewidth]{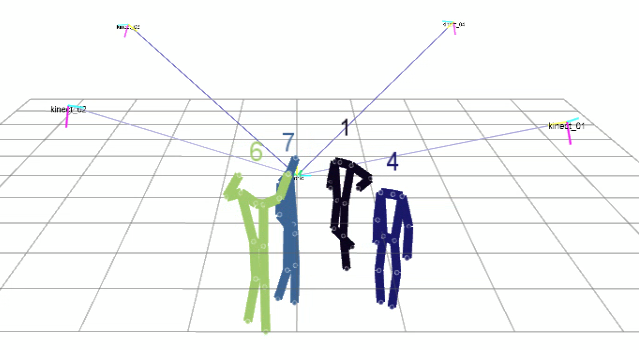}
	\caption{Typical output of the proposed system. Kinematic tracking of four persons at the same time.}
	\label{fig:systemres}
\end{figure}

Fig.~\ref{fig:systemres} shows an example of the virtual scene where the four identified skeletons (the subject ID is presented as the number over the skeleton) replicating the real subject movements.

%The dataset is used to perform a comparison of the proposed approach together with other state-of-the-art approaches.
The evaluation was conducted using as overall metrics the average and standard deviation of the joints displacement tracking errors among all the joints.
%of  difference between   joints 3D positions estimated by the two systems.
%As an overall metric, we consider the mean joint error and its standard deviation, averaged on all the skeleton joints.
In particular, after the interpolation step, for each $m$-th joint, we compare the estimated 3D position and the corresponding ground truth, respectively named $\mathbf{q}_{\text{est,m}}$ and $\mathbf{q}_{\text{gt,m}}$.
Then, given the sequences of $N$ time samples $\{\mathbf{Q}_{\text{est},1},\dots, \mathbf{Q}_{\text{est},N}\}$ and $\{\mathbf{Q}_{\text{gt},1},\dots, \mathbf{Q}_{\text{gt},N}\}$, where $\mathbf{Q}_{\text{est},i} = \sum_{m=1}^{M} \mathbf{q}_{\text{est,m,i}}$ and $\mathbf{Q}_{\text{gt},i} = \sum_{m=1}^{M} \mathbf{q}_{\text{gt,m,i}}$, the two evaluation metrics are defined as:
\begin{equation}
	\begin{aligned}
		e_{\text{avg}} &= \frac{\sum_{i=1}^{N} \| \mathbf{Q}_{\text{est},i}-\mathbf{Q}_{\text{gt},i} \|}{N}, \\
		e_{\text{sd}} &= \sqrt{\frac{\sum_{i=1}^{N} (\| \mathbf{Q}_{\text{est},i}-\mathbf{Q}_{\text{gt},i} \| - e_{\text{avg}})^2}{N}}.
	\end{aligned}
	\label{eq:ave_sd}
\end{equation}

%We report the tracking results, for all the 6 sequences, obtained from:
%
%\begin{itemize} \itemsep0.1em
%	\item[i)] the OpenPose detections reprojected in 3D;
%	\item[ii)] the body poses obtained using a moving average filter, with window size $N=15$, on the OpenPose detections;
%	\item[iii)] the OpenPTrack estimates~\cite{carraro2017real};
%	\item[iv)] the estimates obtained with the proposed solution\footnote{In Section \ref{subsec:example_estimates} we show some examples of the 3D body pose estimates reprojected on the corresponding RGB frames.} (see Section \ref{sec:fusion_algorithm}).
%\end{itemize}

%\begin{figure}
%	\centering
%	\includegraphics[width=\linewidth]{figures/systemresult4people.png}
%	\caption{Typical output of the proposed system. Kinematic tracking of four persons at the same time.}
%	\label{fig:systemres}
%\end{figure}

%Fig.~\ref{fig:systemres} shows an example of the virtual scene where the four identified skeletons (the subject ID is presented as the number over the skeleton) replicating the real subject movements.
%The typical system output is shown in Fig.~\ref{fig:systemres}, where different body configurations are tracked using a network of four Kinects (the numbers close to the target represent the tracking labels). 

For all the six sequences, we computed the same performance metrics (average joint displacement error and standard deviation) on the estimates provided by other state-of-the-art approaches using as input exactly the same data from all the four available Kinects.
The other comparison methods have been: (i) OpenPose~\cite{Cao2017RealtimeM2} enriched with the data association and depth inference algorithms, 
(ii) moving average filtering (MAF), a common baseline approach already described in other similar state-of-the-art works such as~\cite{carraro2017real,carraro2016improved}, and (iii) the standard version of OpenPTrack~\cite{carraro2017real}.
The obtained results are reported in Table~\ref{tab:error_comparison}.

\begin{table}[htbp]
    \caption{Joint displacement tracking errors of the four systems considered for comparison. The proposed approach in bold. Results reported in $cm$ as mean $\pm$ sd.}
	\centering
	%\captionsetup{width=0.95\linewidth}
	\def\arraystretch{1.1}
	\resizebox{\linewidth}{!}{ 
	\begin{tabular}{c|cccc}
		\toprule 
		\textbf{Sequence and} & \multirow{2}{*}{\textbf{OpenPose~\cite{Cao2017RealtimeM2}}} & \multirow{2}{*}{\textbf{MAF}} & \multirow{2}{*}{\textbf{OpenPTrack~\cite{carraro2017real}}} &
		\textbf{Proposed} \\
		\textbf{Subject}  & & & & \textbf{Solution}\\ 
		\midrule
		Seq. 1 & $8.29\pm 7.39$ & $12.56 \pm 7.17$ & $11.85\pm 6.4$ & $\mathbf{5.55\pm 2.68}$\\
		Seq. 2 & $9.47\pm 8.43$ & $17.16\pm 10.23$ & $14.84\pm 8.67$ &  $\mathbf{7.41\pm 4.74}$ \\
		Seq. 3 & $8.79\pm 7.42$ & $14.48\pm 7.35$ & $14.17\pm 6.2$ & $\mathbf{6.74\pm 3.76}$\\
		Seq. 4 & $11.33\pm 8.81$ & $20.68\pm 9.44$ & $18.56\pm 8.24$ &  $\mathbf{9.53\pm 4.46}$\\
		Seq. 5, Sbj. 1 & $8.81\pm 8.74$ & $14.98\pm 7.98$ & $13.17\pm 6.63$ & $\mathbf{6.45\pm 3.52}$\\
		Seq. 5, Sbj. 2 & $9.01\pm9.11$ & $20.73\pm 10.85$ & $17.31\pm 8.86$ & $\mathbf{7.43\pm 5.53}$\\
		Seq. 6, Sbj. 1 & $9.94\pm 9.25$ & $20.93\pm 10.9$ & $17.62\pm 8.91$ & $\mathbf{8.55 \pm 5.1}$\\
		Seq. 6, Sbj. 2 & $9.74\pm 8.71$ & $19.75\pm 9.82$ & $16.11\pm 8.28$ & $\mathbf{8.05\pm 4.31}$ \\
		\bottomrule
	\end{tabular}
	}
\label{tab:error_comparison}
\end{table}

The results clearly show how the proposed multi-subject kinematics tracking approach outperforms, in terms of joint displacement errors, the other considered state-of-the-art systems.
%As it can be seen, the proposed system effectively produces temporally consistent full-3D skeletal poses for the subjects in the scene, clearly outperforming the state-of-the-art systems considered.
Moreover, it is worth noticing that in the fifth and sixth sequences the two subjects on the scene were tracked at the same time, demonstrating the absence of accuracy drops in multi-user applications of our system.

\begin{figure}
	\centering
	%\captionsetup{width=0.6\textwidth}
	\includegraphics[width=0.48\textwidth]{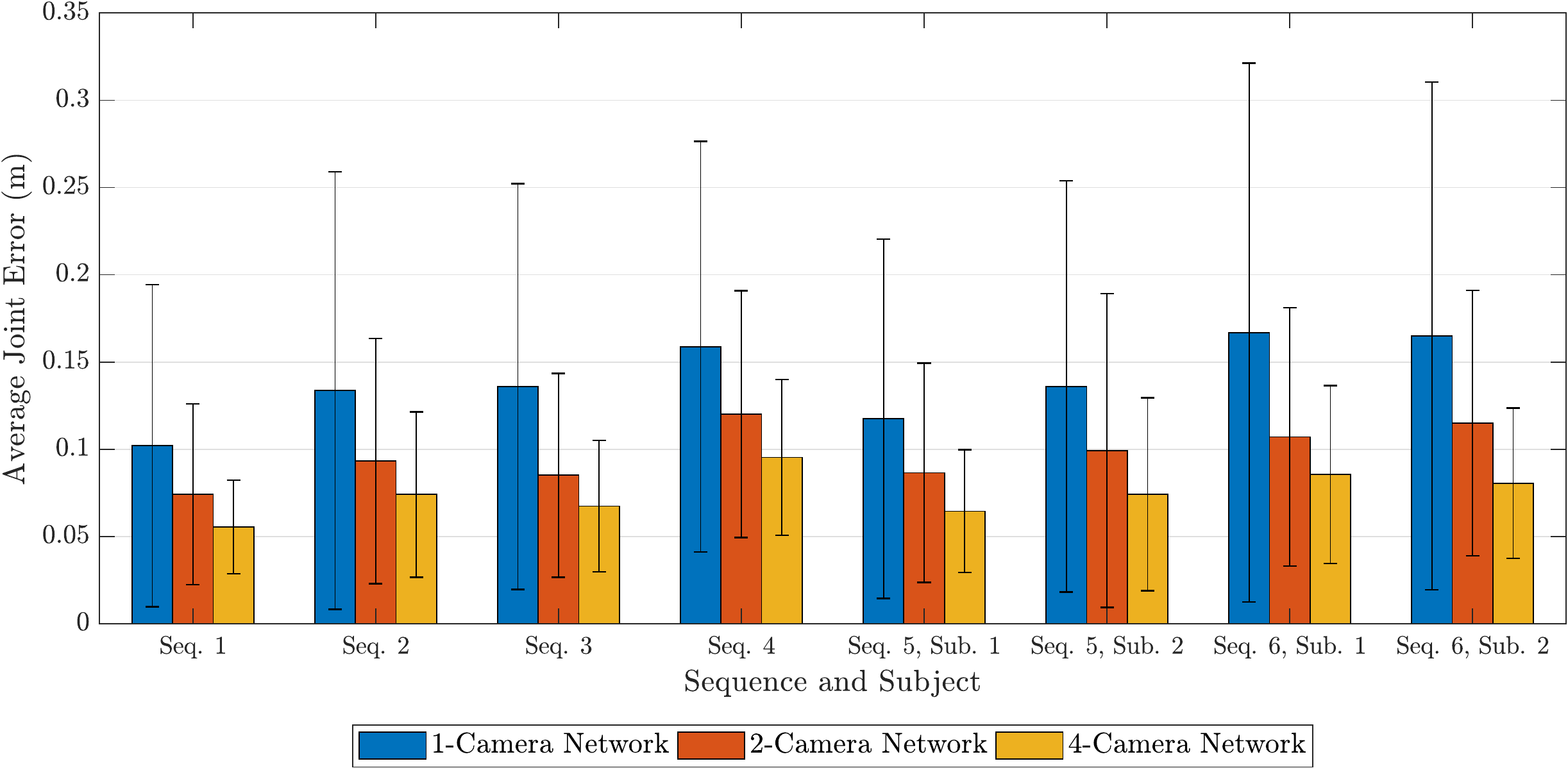}
	\caption{Tracking performance comparison, in terms of mean joint error and standard deviation averaged on all joints, using networks of different sizes. The thin bars in the graph indicate an interval of one standard deviation around the error mean.}
	\label{fig:camera_network}
\end{figure}

As an additional evaluation metric for the proposed system, we investigated the consistence of the estimates quality when the network is composed by a lower number of cameras. To do that we selected, for every available sequence, data coming just from two, three, or four \textit{detection nodes}.
Fig.~\ref{fig:camera_network} shows how the estimation quality increases with the number of cameras installed in the network.\\
%
%The single-view body pose detection system we use in this thesis is working at $\sim 34$-$35\,\text{Hz}$.
%We tested the proposed tracking algorithm using an Intel Core i7-4770 CPU.
%The tracking algorithm works at $\sim 100\, \text{Hz}$, being that it requires $\sim0.01 \,\text{s}$ ($\sim 0.015 \,\text{s}$ in the worst case scenario) to update the skeleton track\footnote{It is worth to note that the results we are referring to are the ones obtained with a version of the code that has not been profiled yet. We believe that it is possible to improve the performance by profiling and by using more efficient libraries.}, given the new detection\footnote{In this section, the considerations about the computational time required by the tracking procedure are all referred to the scenario with only one subject present in the scene.}.
%The time required for the execution of each part of the algorithm, in order to understand the relation with the overall computation time. 
Furthermore, it is important to observe that such performances were obtained while producing the body pose estimates in real-time (tested on an Intel Core i7-4770 CPU and Nvidia GeForce GTX 1060 GPU powered desktop).
In a typical iteration, the average computational time (approx. $13.4 \, ms$) is distributed among the different major processes as follow:
\begin{itemize} \itemsep 0.1em
	\item $\sim 40\,\mu s$ for the adaptive scheme with the outlier detection, the iterative computation of the outlier thresholds, and the iterative estimate of the limb lengths;
	\item $\sim 3$ -- $5 \, ms$ for the limb lengths optimization;
	\item $\sim 5$ -- $7 \, ms$ for the update step of the Kalman filter;
	\item $\sim 1 \, ms$ for other computational activities.
\end{itemize}
%The implementation of the adaptive scheme and the track parameters update step are negligible with respect to the total computation time.
%
\begin{table}
    \caption{Theoretical and real computational times of our systems depending on the number of subjects to track.}
    %\tiny
    %\small
    \centering
    %\captionsetup{width=0.95\linewidth}
    \def\arraystretch{1.1}
    \resizebox{\linewidth}{!}{ 
        \begin{tabular}{c|ccc}
            \toprule 
            \textbf{Tracked} & \textbf{Worst case comp. time} & \textbf{Worst case fps} & \textbf{Experimental fps} \\
            \textbf{persons}  & $[ms]$ & \textbf{(theoretical)} &  \\
            \midrule
            1 & $ \sim13.4 $ & $ \sim74.6 $ &$ 33.9 $ \\
            2 & $ \sim26.8 $ & $ \sim37.3 $ &$ 32.1 $ \\
            3 & $ \sim40.2 $ & $ \sim24.9 $ & --  \\
            4 & $ \sim53.6 $ & $ \sim18.7 $ &$ 18.7 $ \\
            5 & $ \sim67.0 $ & $ \sim14.9 $ & --  \\
            6 & $ \sim80.4 $ & $ \sim12.4 $ & --  \\
            \bottomrule
        \end{tabular}
    }
\label{tab:comp_time}
\end{table}
Table~\ref{tab:comp_time} reports how the theoretical output frame rate scales with the number of subjects to be tracked by our system.
In case of up to two subjects the actual tracking output frame rate is however lower than the theoretical one due to the input data frequency constraint; indeed, the Kinect cameras are capable of providing data at maximum at $30\, fps$\footnote{Due to the lack of constraints in cameras synchronization placed by our system, however, it is feasible to achieve slightly higher data input frame rates.}.
%
%Given the maximum frame rate of the Kinect camera ($30\, fps$), the 
%Standing the limits in the maximum input data frame rate ($30\, fps$)
%The complete camera network setup (4 Kinects, 4 detection nodes, 1 computational node) is capable of tracking the movements of one subject at $33.9\,fps$, and at $32.1\,fps$ when tracking two persons.
%In such case the total execution time is lower than the time between two consecutive frames, therefore the performance of the system is defined by the Kinects framerate.
In case of three or more persons, instead, the most restrictive limit becomes the computational time, with an actual tracking frame rate which scales proportionally to the number of subjects to track.
Setting a limit for considering an application as real-time to be equal to $15\, fps$, it is shown that for cases of up to 5 persons in the scene our approach satisfies this constraint.

\addtolength{\textheight}{-1.5cm}   % This command serves to balance the column lengths

\section{Conclusions}
\label{sec:conclusions}

In this work we proposed, discussed, and evaluated an innovative markerless approach to accurately estimate and the track human motion in real-time through a calibrated network of RGB-D cameras.
The proposed methodology demonstrated to be reliable and accurate in tracking multiple persons at the same time, without requiring the subjects to perform initial calibration activities or to wear any marker.
The developed system requires neither a specific number of cameras in the network, nor all the cameras to be of the same manufacturer; moreover, it does not require the cameras to be synchronized. These three valuable features enable users to build their own heterogeneous network following their specific needs and possibilities.

Starting from the noisy single-frame single-view detections, our algorithm ensures temporal and physical consistency during the whole tracking period.
As demonstrated by the obtained results, it reduces the joint displacement tracking error, with respect to the current state-of-the-art approaches used as reference for the comparison.
%Furthermore, the obtained results demonstrates a significant reduction 
%In this way, we can observe a notable improvement 
%of both average joint tracking error and standard deviation, with respect to the original single-view 3D detections and to other state-of-the-art works used as a baseline.
%The experimental results also show how the tracking performance of the system worsen during very rapid movements.
%Indeed, the end joints, which perform the fastest movements in a human body, present the poorest performance.
%Their motion is therefore difficult to predict due to the constant velocity model employed in the Kalman filter implementation.
Moreover, the proposed approach is not only reliable and effective, but also efficient, enabling to track in real-time up to 5 subjects at the same time.
%We also demonstrated that the proposed solution is not just effective, but also efficient, given the low computational burden required.
This is indeed a crucial capability, since it is a major requirement for the large majority of the applications in the human-robot interaction field.
%the large majority of our targeted applications requires a real-time tracking of human movements.
%Indeed, the ability to track the human body pose in real-time is a fundamental characteristic of our system that is of paramount importance for many applications.

%In order to obtain a further performance improvement, we think that the focus should be put on the single-view skeleton detections. 
%A first interesting idea could be to implement an outlier detection procedure based on the new predicted positions of the skeleton joints, along with their innovation covariance.
%In this way, even in fast sequences, we could try and detect possible outlier measurements and obtain a more stable track.
%
%Moreover, we could include a tracking procedure directly on the 2D estimates of the single cameras, considering the joint confidence given as output by the OpenPose detector.
%
%Finally, we think that taking into account the relative distance between camera and tracked subject could be an interesting idea to consider.
%Indeed, using an adaptive scheme, we could give more importance to the measurements coming from a sensor closer to the subject, being that we can expect its depth information to be more reliable.

%%%%%%%%%%%%%%%%%%%%%%%%%%%%%%%%%%%%%%%%%%%%%%%%%%%%%%%%%%%%%%%%%%%%%%%%%%%%%%%%

%%%%%%%%%%%%%%%%%%%%%%%%%%%%%%%%%%%%%%%%%%%%%%%%%%%%%%%%%%%%%%%%%%%%%%%%%%%%%%%%

%%%%%%%%%%%%%%%%%%%%%%%%%%%%%%%%%%%%%%%%%%%%%%%%%%%%%%%%%%%%%%%%%%%%%%%%%%%%%%%%
\section*{Acknowledgment}
Authors would like to thank Dr. Zimi Sawacha and the Magick team for their help in collecting and processing the ground truth references for our database with their marker-based motion capture system.

%%%%%%%%%%%%%%%%%%%%%%%%%%%%%%%%%%%%%%%%%%%%%%%%%%%%%%%%%%%%%%%%%%%%%%%%%%%%%%%%
\bibliographystyle{IEEEtran}
\bibliography{bibliography}

\begin{thebibliography}{10}
\providecommand{\url}[1]{#1}
\csname url@samestyle\endcsname
\providecommand{\newblock}{\relax}
\providecommand{\bibinfo}[2]{#2}
\providecommand{\BIBentrySTDinterwordspacing}{\spaceskip=0pt\relax}
\providecommand{\BIBentryALTinterwordstretchfactor}{4}
\providecommand{\BIBentryALTinterwordspacing}{\spaceskip=\fontdimen2\font plus
\BIBentryALTinterwordstretchfactor\fontdimen3\font minus
  \fontdimen4\font\relax}
\providecommand{\BIBforeignlanguage}[2]{{%
\expandafter\ifx\csname l@#1\endcsname\relax
\typeout{** WARNING: IEEEtran.bst: No hyphenation pattern has been}%
\typeout{** loaded for the language `#1'. Using the pattern for}%
\typeout{** the default language instead.}%
\else
\language=\csname l@#1\endcsname
\fi
#2}}
\providecommand{\BIBdecl}{\relax}
\BIBdecl

\bibitem{carraro2016powerful}
M.~Carraro, M.~Munaro, and E.~Menegatti, ``A powerful and cost-efficient human
  perception system for camera networks and mobile robotics,'' in
  \emph{International Conference on Intelligent Autonomous Systems}.\hskip 1em
  plus 0.5em minus 0.4em\relax Springer, 2016, pp. 485--497.

\bibitem{sarkka2015adaptive}
S.~S{\"a}rkk{\"a}, V.~Tolvanen, J.~Kannala, and E.~Rahtu, ``Adaptive kalman
  filtering and smoothing for gravitation tracking in mobile systems,'' in
  \emph{Indoor Positioning and Indoor Navigation (IPIN), 2015 International
  Conference on}.\hskip 1em plus 0.5em minus 0.4em\relax IEEE, 2015, pp. 1--7.

\bibitem{carraro2017real}
M.~Carraro, M.~Munaro, J.~Burke, and E.~Menegatti, ``Real-time marker-less
  multi-person 3d pose estimation in rgb-depth camera networks,'' \emph{arXiv
  preprint arXiv:1710.06235}, 2017.

\bibitem{sarafianos20163d}
N.~Sarafianos, B.~Boteanu, B.~Ionescu, and I.~A. Kakadiaris, ``3d human pose
  estimation: A review of the literature and analysis of covariates,''
  \emph{Computer Vision and Image Understanding}, vol. 152, pp. 1--20, 2016.

\bibitem{han2017simultaneous}
F.~Han, X.~Yang, C.~Reardon, Y.~Zhang, and H.~Zhang, ``Simultaneous feature and
  body-part learning for real-time robot awareness of human behaviors,'' in
  \emph{IEEE International Conference on Robotics and Automation (ICRA)}, 2017,
  pp. 2621--2628.

\bibitem{zanfir2013moving}
M.~Zanfir, M.~Leordeanu, and C.~Sminchisescu, ``The moving pose: An efficient
  3d kinematics descriptor for low-latency action recognition and detection,''
  in \emph{Proceedings of the IEEE International Conference on Computer
  Vision}, 2013, pp. 2752--2759.

\bibitem{Ghidoni:2017:MFR:3065975.3066116}
\BIBentryALTinterwordspacing
S.~Ghidoni and M.~Munaro, ``A multi-viewpoint feature-based re-identification
  system driven by skeleton keypoints,'' \emph{Robot. Auton. Syst.}, vol.~90,
  no.~C, pp. 45--54, Apr. 2017. [Online]. Available:
  \url{https://doi.org/10.1016/j.robot.2016.10.006}
\BIBentrySTDinterwordspacing

\bibitem{jaimes2007multimodal}
A.~Jaimes and N.~Sebe, ``Multimodal human--computer interaction: A survey,''
  \emph{Computer vision and image understanding}, vol. 108, no.~1, pp.
  116--134, 2007.

\bibitem{mccoll2011human}
D.~McColl, Z.~Zhang, and G.~Nejat, ``Human body pose interpretation and
  classification for social human-robot interaction,'' \emph{International
  Journal of Social Robotics}, vol.~3, no.~3, pp. 313--332, 2011.

\bibitem{gupta20113d}
A.~Gupta, S.~Satkin, A.~A. Efros, and M.~Hebert, ``From 3d scene geometry to
  human workspace,'' in \emph{Computer Vision and Pattern Recognition (CVPR),
  2011 IEEE Conference on}.\hskip 1em plus 0.5em minus 0.4em\relax IEEE, 2011,
  pp. 1961--1968.

\bibitem{morato2014toward}
C.~Morato, K.~N. Kaipa, B.~Zhao, and S.~K. Gupta, ``Toward safe human robot
  collaboration by using multiple kinects based real-time human tracking,''
  \emph{Journal of Computing and Information Science in Engineering}, vol.~14,
  no.~1, p. 011006, 2014.

\bibitem{chen2012we}
C.~Chen and J.-M. Odobez, ``We are not contortionists: Coupled adaptive
  learning for head and body orientation estimation in surveillance video,'' in
  \emph{Computer Vision and Pattern Recognition (CVPR), 2012 IEEE Conference
  on}.\hskip 1em plus 0.5em minus 0.4em\relax IEEE, 2012, pp. 1544--1551.

\bibitem{chen2011combined}
C.~Chen, A.~Heili, and J.-M. Odobez, ``Combined estimation of location and body
  pose in surveillance video,'' in \emph{Advanced Video and Signal-Based
  Surveillance (AVSS), 2011 8th IEEE International Conference on}.\hskip 1em
  plus 0.5em minus 0.4em\relax IEEE, 2011, pp. 5--10.

\bibitem{Cao2017RealtimeM2}
Z.~Cao, T.~Simon, S.-E. Wei, and Y.~Sheikh, ``Realtime multi-person 2d pose
  estimation using part affinity fields,'' \emph{2017 IEEE Conference on
  Computer Vision and Pattern Recognition (CVPR)}, pp. 1302--1310, 2017.

\bibitem{mehta2017vnect}
D.~Mehta, S.~Sridhar, O.~Sotnychenko, H.~Rhodin, M.~Shafiei, H.-P. Seidel,
  W.~Xu, D.~Casas, and C.~Theobalt, ``Vnect: Real-time 3d human pose estimation
  with a single rgb camera,'' \emph{ACM Transactions on Graphics (TOG)},
  vol.~36, no.~4, p.~44, 2017.

\bibitem{zivkovic2010wireless}
Z.~Zivkovic, ``Wireless smart camera network for real-time human 3d pose
  reconstruction,'' \emph{Computer Vision and Image Understanding}, vol. 114,
  no.~11, pp. 1215--1222, 2010.

\bibitem{carraro2016cost}
M.~Carraro, M.~Munaro, and E.~Menegatti, ``Cost-efficient rgb-d smart camera
  for people detection and tracking,'' \emph{Journal of Electronic Imaging},
  vol.~25, pp. 041\,007--041\,007, 04 2016.

\bibitem{buys2014adaptable}
K.~Buys, C.~Cagniart, A.~Baksheev, T.~De~Laet, J.~De~Schutter, and
  C.~Pantofaru, ``An adaptable system for rgb-d based human body detection and
  pose estimation,'' \emph{Journal of visual communication and image
  representation}, vol.~25, no.~1, pp. 39--52, 2014.

\bibitem{carraro2016improved}
M.~Carraro, M.~Munaro, A.~Roitberg, and E.~Menegatti, ``Improved skeleton
  estimation by means of depth data fusion from multiple depth cameras,'' in
  \emph{International Conference on Intelligent Autonomous Systems}.\hskip 1em
  plus 0.5em minus 0.4em\relax Springer, 2016, pp. 1155--1167.

\bibitem{moon2016multiple}
S.~Moon, Y.~Park, D.~W. Ko, and I.~H. Suh, ``Multiple kinect sensor fusion for
  human skeleton tracking using kalman filtering,'' \emph{International Journal
  of Advanced Robotic Systems}, vol.~13, no.~2, p.~65, 2016.

\bibitem{caon2011context}
M.~Caon, Y.~Yue, J.~Tscherrig, E.~Mugellini, and O.~A. Khaled, ``Context-aware
  3d gesture interaction based on multiple kinects,'' in \emph{Proceedings of
  the first international conference on ambient computing, applications,
  services and technologies, AMBIENT}.\hskip 1em plus 0.5em minus 0.4em\relax
  Citeseer, 2011, pp. 7--12.

\bibitem{kaenchan2013automatic}
S.~Kaenchan, P.~Mongkolnam, B.~Watanapa, and S.~Sathienpong, ``Automatic
  multiple kinect cameras setting for simple walking posture analysis,'' in
  \emph{Computer Science and Engineering Conference (ICSEC), 2013
  International}.\hskip 1em plus 0.5em minus 0.4em\relax IEEE, 2013, pp.
  245--249.

\bibitem{liu2018human}
G.~Liu, G.~Tian, J.~Li, X.~Zhu, and Z.~Wang, ``Human action recognition using a
  distributed rgb-depth camera network,'' \emph{IEEE Sensors Journal}, vol.~18,
  no.~18, pp. 7570--7576, 2018.

\bibitem{kamal2016distributed}
A.~T. Kamal, J.~H. Bappy, J.~A. Farrell, and A.~K. Roy-Chowdhury, ``Distributed
  multi-target tracking and data association in vision networks,'' \emph{IEEE
  transactions on pattern analysis and machine intelligence}, vol.~38, no.~7,
  pp. 1397--1410, 2016.

\bibitem{ershadi2018multiple}
S.~Ershadi-Nasab, E.~Noury, S.~Kasaei, and E.~Sanaei, ``Multiple human 3d pose
  estimation from multiview images,'' \emph{Multimedia Tools and Applications},
  vol.~77, no.~12, pp. 15\,573--15\,601, 2018.

\bibitem{munaro2016openptrack}
M.~Munaro, F.~Basso, and E.~Menegatti, ``Openptrack: Open source multi-camera
  calibration and people tracking for rgb-d camera networks,'' \emph{Robotics
  and Autonomous Systems}, vol.~75, pp. 525--538, 2016.

\bibitem{kuhn1955hungarian}
H.~W. Kuhn, ``The hungarian method for the assignment problem,'' \emph{Naval
  Research Logistics (NRL)}, vol.~2, no. 1-2, pp. 83--97, 1955.

\bibitem{munkres1957algorithms}
J.~Munkres, ``Algorithms for the assignment and transportation problems,''
  \emph{Journal of the society for industrial and applied mathematics}, vol.~5,
  no.~1, pp. 32--38, 1957.

\end{thebibliography}

\end{document}